# Convolutional Neural Networks based Intra Prediction for HEVC

Wenxue Cui, Tao Zhang, Shengping Zhang,

Feng Jiang, Wangmeng Zuo and Debin Zhao

*School of Computer Science*

*Harbin Institute of Technology, Harbin, 150001, China*
wenxuecui@stu.hit.edu.cn;  {taozhang.cs, s.zhang, fjiang}@hit.edu.cn

*Abstract*: Traditional intra prediction methods for HEVC rely on using the nearest reference lines for predicting a block, which ignore much richer context between the current block and its neighboring blocks and therefore cause inaccurate prediction especially when weak spatial correlation exists between the current block and the reference lines. To overcome this problem, in this paper, an intra prediction convolutional neural network (IPCNN) is proposed for intra prediction, which exploits the rich context of the current block and therefore is capable of improving the accuracy of predicting the current block. Meanwhile, the predictions of the three nearest blocks can also be refined. To the best of our knowledge, this is the first paper that directly applies CNNs to intra prediction for HEVC. Experimental results validate the effectiveness of applying CNNs to intra prediction and achieved significant performance improvement compared to traditional intra prediction methods.

## 1. Introduction

The latest video coding standard H.265/HEVC [1] developed by the Joint Collaborative Team on Video Coding (JCT-VC) provides a similar perceptual quality but with about 50% bitrate saving compared to its predecessor H.264/AVC because of the introduction of many advanced technologies. For instance, the quadtree-based coding unit data structure is used and the number of angular direction is extended to 33 in HEVC. The fine-grained direction can provide more accurate prediction in HEVC. HEVC also utilizes hybrid video coding framework. But a lot of improvements have been made by contrast with H.264/AVC. In H.264/AVC, the coding unit is a macro block with a fixed size of 16x16. In contrast, HEVC introduces the conceptions of coding tree unit (CTU), prediction unit (PU), translation unit (TU), and coding unit (CU). An image is initially partitioned into CTUs. The size of CTU can be 16x16, 32x32 or 64x64. For a CTU, it can be coded as a whole, or to be split into four CU evenly. CU is one of the most basic module for inter coding and intra coding whose size can be from 8x8 to 64x64. It can also be divided into four parts evenly according to the texture of an image. PU is one of the most basic unit for intra prediction which is separated from CU and its size can be from 4x4 to 64x64. TU is used for transformation and quantization. All TUs in a PU will share the same prediction information. The size of a TU varies from 4x4 to 32x32. Due to the strong spatial correlation of images, a small TU can reduce the prediction error between prediction pixels and reference pixels.

In the intra prediction of HEVC, the generation of angular prediction follows the minimum error between prediction pixels and original values. However, this kind of methods may not generate the optimal prediction because it only assumes that the texture information follow a pure direction. Therefore, noises or occlusion on the nearest reference line may lead to a large prediction deviation. To solve this problem, a lot of improvements have been proposed. In [2], a template matching method is proposed, in which the template of the coding block is searched in the non-local reference content and the corresponding block with the best matched template is selected as the content. This method performs well at the area with many duplicated patterns. However, it has high computational complexity as there is a complicated matching procedure. A position-dependent filtering method based on MMSE is proposed in [3] where each position has its own weights according to its block size or directions. However, a tremendous extra weight information is needed when decoding. In order to avoid this obstacle, in [4-7], markov process based methods are presented which refine the intra prediction by modeling according to the correlation between adjacent pixels. The image inpainting is another widely used technology for improving intra prediction. An edge-based inpainting algorithm is proposed in [8-9] . The partial differential equation based on image inpainting is also used in [10-11]. Most of the aforementioned methods for HEVC rely on using the nearest reference lines for predicting a block, which ignore much richer context between the current block and its neighboring blocks and therefore cause inaccurate prediction especially when weak spatial correlation exists between the current block and the reference lines.

To overcome the shortcomings of traditional methods, in this paper, we proposed a convolutional neural networks (CNNs) based intra prediction method which uses more context information to predict the current PU block. Specifically, an intra prediction CNN (IPCNN) is presented for intra prediction. This network can learn the spatial correlation between current PU block and its nearest reference blocks. The proposed method has several appealing properties. First, it could be embedded into HEVC directly without too much modification and provides high prediction accuracy comparing with the traditional method in HEVC. Second, the prediction of the nearest three reconstruction blocks can also be improved in the prediction process.

Overall, the contributions of this work are three-fold:

1) We devise an idea to perform intra prediction for HEVC, which exploits the rich context between the current block to be predicted and its three nearest blocks. This idea is significant different from traditional intra prediction methods which only relies on using the nearest reference lines for predicting the current block.

2) We propose a convolutional neural networks (CNNs) based method for intra prediction, which exploit the context of the current block to improve its prediction accuracy while refining the prediction of its three nearest blocks.

3) Our experimental results validate the effectiveness of applying CNNs to intra prediction for HEVC, which provides new insights of applying CNNs to improve the performance of intra coding to researchers in this field.

The rest of the paper is organized as follows. Section 2 provides a brief overview of related work. Section 3 describes the proposed method. Experimental results are presented in Section 4. More discussions are given in Section 5. Finally, Section 6 concludes this paper.

## 2. Related Work

### 2.1 The Intra Prediction in HEVC

The intra prediction coding in HEVC follows a block-wise approach. Specifically, HEVC partitions each frame hierarchically into non-overlapping coding units (CU) and further into prediction units (PU). Then each PU will be predicted according to the nearest reference pixels and the best prediction mode within the same frame. For each PU to be coded, prediction blocks for the luminance and chrominance are created by the reconstructed pixels surrounding the current PU block. The quality of the prediction will have a direct influence on the bitrate saving and PSNR of video sequences. After prediction coding, the residual signal will be achieved which is the difference between the original block and its prediction values. Transform and quantization process will be implemented continuously. Ultimately, the quantized coefficients will be fed to the entropy coding process directly.

In HEVC, a set of 35 modes including 33 angular modes and a DC and a PLANAR mode are available. In intra prediction, the set of bounding pixels are used to generate the block's estimate.

### 2.2 Convolutional Neural Network based Video Coding

Recently, The Convolutional neutral network (CNN) is used universally in many domains. Many trials has been done by using CNN for super-resolution (SR) [12] image classification [13] and object detection [14]. The CNN is well known to perform well for image processing and classification. However, the CNN is rarely used in video coding. In [15-17], CNN is used for Intra CU mode decision. In these papers, the input of the network is a 8x8 block and two classes are extracted at the last layer. In fact, it is a classification problem. The result of the network is used for decision that if this input block will be split or not. CNN is also used in in-loop filtering for coding efficiency improvement in [18]. Besides, [19] proposed a CNN-based approach for post-processing in HEVC intra coding. It is from AR-CNN which is used for compression artifacts reduction. However, these methods mentioned above are not used for intra prediction directly.

## 3. Proposed Method

In this section, the proposed method will be given in detail. First, the framework is described. Second, the datasets for training are discussed comprehensively. Last, the structure of the proposed network and the details for training are shown.

## 3.1 Overview

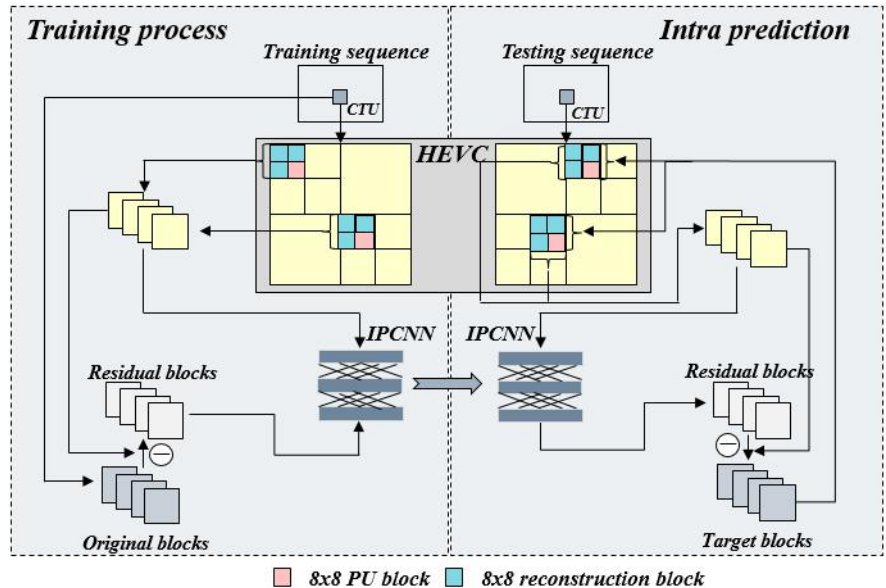

Fig. 1 The framework of the proposed method

The proposed method involves two processes: training and intra prediction as shown in Fig. 1. In the training process, a set of training samples are extracted from HEVC intra coding which consists of the intra prediction block (from HEVC) for the 8x8 PU (pink block) and its three nearest 8x8 reconstruction blocks (green blocks). For more details of extracting training samples, the next sub-section will be discussed. The output of the network (IPCNN) is the residual blocks generated by subtracting the original blocks from the input block. After the training process, a complete network is achieved. This network will be used in the intra prediction process.

In the intra prediction process, a 16x16 context block containing the best intra prediction from HEVC for the 8x8 PU block and its three nearest reconstruction blocks is fed to the network learned in the training process. The residual block will be generated as the output of the network. Then, the target block will be obtained by subtracting the residual block from the input block. At last, the predicted 8x8 PU block in the 16x16 target block will be sent to the HEVC. The original 8x8 PU block in HEVC will be replaced by the target PU block. The remaining processes are the same with HEVC. The residual signal which is the difference between the original block and its prediction values will be achieved by subtracting prediction blocks from the original blocks. Transform and quantization process will be implemented subsequently. Finally, the quantized coefficients will be directly fed to the entropy coding process.

## 3.2 Datasets for Training

In order to achieve better performance, the datasets for training are elaborated. Two vital properties of the training data are discussed below. One is the size of the training data. The appropriateness of the data size will make the performance of CNN better. The other is interdependency of the training data which reflects the quantity of information that the training data carries.

In HEVC, the minimum size of a PU block is 4x4, which is too small to be used as the training data. Because the information it carries is not sufficient. The second-smallest size of a PU block is 8x8. It is better than 4x4 discussed above. However, the prediction of the current PU is generated from the top reference pixels and left reference pixels. A portion of information has lost because of deficiency of the nearest reference pixels. So, the indispensability of the nearest reference pixels is obvious. Conversely, if the size of the PU block and reference pixels are larger, the computational complexity will be higher. Ultimately, for the input data of CNN, this paper selects the size of 16x16 which includes a 8x8 PU block and its three nearest three reconstruction 8x8 blocks as shown in Fig. 2. The 8x8 PU block (red area of Fig. 2) is the prediction by using the best intra mode in HEVC according to reference lines. For the output data, it is also a 16x16 block which is the residual between the input data and original data. In summary, there are three requirements: a) the size of the current PU block is 8x8 in proposed method; b) the initial values of the PU block are predicted by the best mode in HEVC; c) the reconstruction blocks are the references which are used for prediction of the PU block.

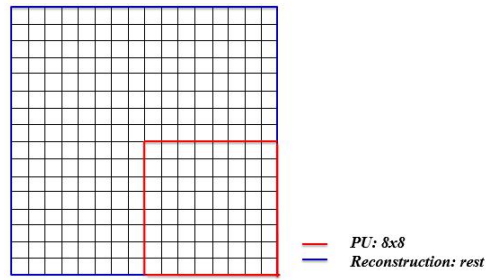

Fig. 2 The shape of training data

The size of 16x16 is chosen for prediction with the following three reasons:

1) It contains the current PU block and three nearest reference blocks which will provide more information to the network for predicting. Therefore, it is beneficial for learning the model.

2) The patterns in the PU block will be learned by the network. It will provide an evidence for the accuracy of prediction block in HEVC.

3) The upper left three reference blocks provide the reconstruction patterns for the network in HEVC. It is possible to make reconstruction processing more accurate.

### 3.3 Details of Our Network Structure and Training

In fact, both the reconstruction error and the prediction error can be regarded as noises in an image. Therefore, the input image can be considered as a noisy observation $Y = X + V$ where X represents the original image and V the noises. The network aims to learn a mapping function $F(Y) = X$ to predict the original image. Instead of directly learning the mapping F, the residual mapping $R(Y) \approx V$ will be learned. Finally, we get the estimated original image by $X \approx Y - R(Y)$.

In this section, we first introduce the architecture of the intra prediction CNN (IPCNN) as shown in Fig. 3. IPCNN is composed of 10 weight layers. There are three types of layers in the IPCNN architecture: Conv + ReLU, Conv + BN + ReLU and Conv. (i) Conv

+ ReLU: for the first layer, 64 filters of size 3×3×c[1] are used to generate 64 feature maps. (ii): Conv + BN + ReLU: for layers 2 ~ 9, 64 filters of size 3×3×64 are used, and batch normalization is added between convolution and ReLU. (iii): Conv: for the last layer, c filters of size 3x3x64 are used to reconstruct the output. Residual learning and batch normalization are applied to speed up the training process and boost the performance.

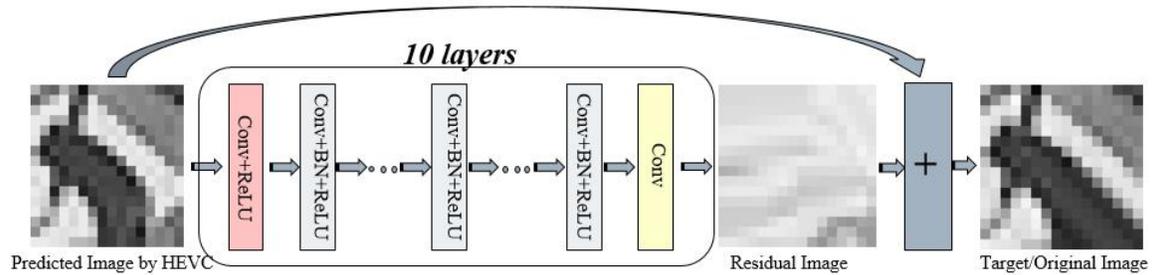

Fig. 3 the network (IPCNN) used in the proposed method

When training the network, some empirical settings are beneficial to achieve better performance. First, the learning rate should be limited in a small range. Maybe it is because the difference between predicted PU block and original block is large which will cause the instability of the network if the learning rate is too large. Second, the batch size should be reduced gradually during the training process. The batch size is reduced by half every 10 epochs from 128 to 32 in this paper.

## 4. Experimental Results

In this section, the experiments are performed to verify the effectiveness of the proposed method. Besides, some analyses are given to show the patterns that our network has learned.

### 4.1 Experimental Settings

In the experiment, the PU block (the red block in Fig. 2) of the training data is used for intra prediction in HEVC. Besides, we use the deep learning framework MatConvnet for training our network. It is easy to embed the learned network into the HEVC reference software. To verify the performance of the proposed scheme, we implement it into HM-14.0. The test sequences include a large range of HEVC standard test video sequences. Our training datasets are from 10 sequences of four different quantization parameters (QPs): 22, 27, 32 and 37, and only luminance component is considered. For each QP, a separate network is trained. While comparing with HEVC, the results are evaluated by BD-Rate [20], where the negative number indicates bitrate saving and the positive number indicates bitrate increase. After comparison by the BD-Rate results, we also analyze the patterns that the network has learned and show some figures for details.

---

[1] c represents the number of image channels

## 4.2 Experimental Results

The experimental results of the proposed method which based on CNN are shown in Table 1. The average BD-rate saving is 0.70%. The maximum bitrate saving is about 1.2% for the sequence PeopleOnStreet.

Table 1 BD-rate saving for The Proposed Scheme with Ranges of Sequences

| Sequences | BD-rate | Sequences | BD-rate |
| --- | --- | --- | --- |
| Traffic | -0.9% | PartyScene | -0.5% |
| PeopleOnStreet | -1.2% | RaceHorses | -0.7% |
| Kimono | -0.2% | BasketballPass | -0.4% |
| ParkScene | -0.8% | BQSquare | -0.1% |
| Cactus | -0.8% | BlowingBubbles | -0.7% |
| BasketballDrive | -0.6% | RaceHorses | -0.7% |
| BQTerrace | -0.8% | FourPeople | -0.3% |
| BasketballDrill | -0.5% | Johnny | -1.0% |
| BQMall | -0.6% | KristenAndSara | -0.8% |
| All average | -0.70% | | |

## 4.3 More Analysis

The CNN can learn the latent patterns between input data and original data. In order to show the patterns that CNN has learned visually, some figures are given in Fig. 4

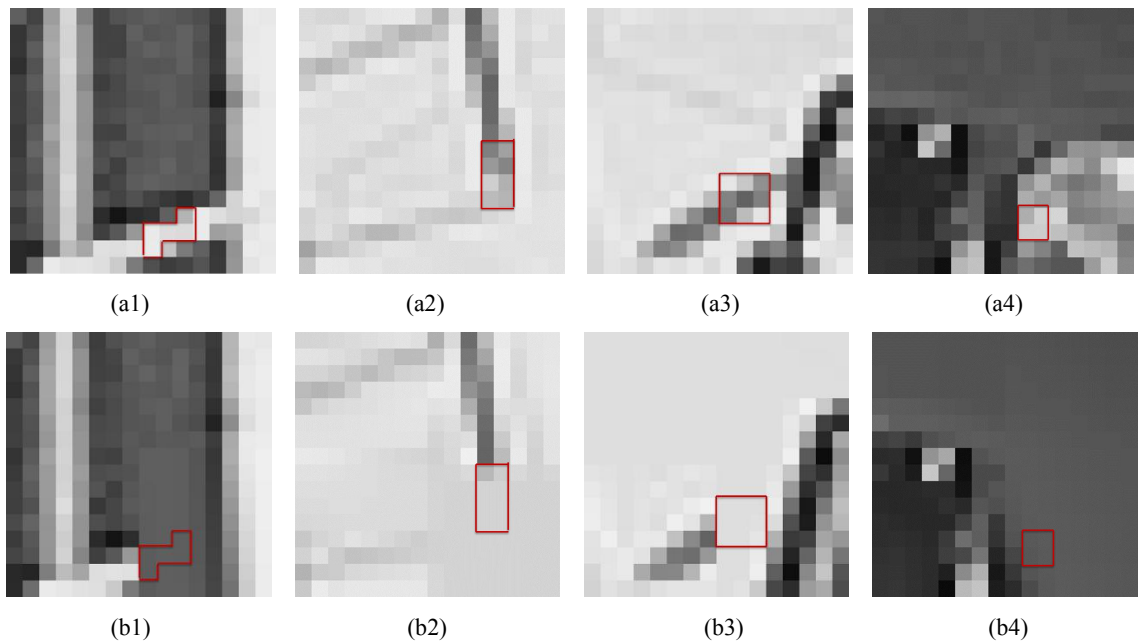

(a1)　　(a2)　　(a3)　　(a4)

(b1)　　(b2)　　(b3)　　(b4)

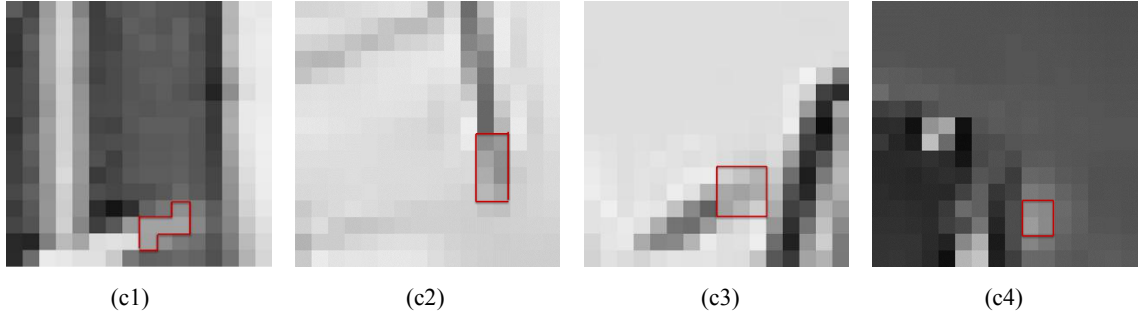

|  (c1)  |  (c2)  |  (c3)  |  (c4)  |

Fig. 4 the results of 4 sets for images (16x16) by columns. (ai) are the original image, (bi) are the prediction blocks from HEVC and (ci) are the prediction result with our method. (i = 1, 2, 3, 4 in "ai", "bi", "ci") and the red area is the region of interest in images.

We can get some knowledges from the results in Fig. 4. In the first set of images (a1, b1, c1), a lighter color is produced in the region of interest. In the second and third set of images, the blank pixels is filled up following a certain rule in the region of interest. From the sets of images discussed above we can find the extension of color from upper left reference blocks to the PU block. The prediction patterns of these three sets of images are obvious visually. However, for the fourth set of images, the pattern that learned by CNN is unobvious, which is the statistic characteristic from the training data.

## 5. Discussion

The 16x16 context block contains one 8x8 PU and three nearest 8x8 reconstruction blocks. In this paper, the 8x8 PU is used for intra prediction. This proposed method in this paper is just a preliminary work. Some expanded ideas are discussed below.

1) Similar to the 8x8 PU, the three nearest 8x8 reconstruction blocks also can obtain an improvement for different QPs. It is shown in Table 2. In this table, 5000 16x16 blocks are tested for 4 different QPs. But only the upper left three reconstruction blocks were included in the calculation of MSE. Specifically, the original MSE in Table 2 represents the MSE between original blocks and reconstructed blocks in HEVC. The target MSE represents MSE between the original blocks and the refined blocks achieved by using proposed method. In Table 2, the target MSE is smaller than the original MSE. Therefore, these nearest reconstruction blocks obtain quality improvement at different QPs. This improvement can be used for refining the reconstruction blocks in HEVC.

2) In this paper, the size of current PU and reconstruction block is 8x8. Maybe it is worth trying the size of 16x16 or larger. When the size of reconstruction block become larger, more reference pixels can be utilized.

3) In the proposed method, the prediction for current PU block maybe not as accurate as the original prediction by HEVC for some block. In order to separate them, a binary classifier can be a good solution. Before using the proposed method, the characteristics of the PU can be analyzed to know if the PU can be predicted well by CNN-based method. If this binary classifier works well, more coding gains will be achieved.

4) In order to enhance the correlation between current PU block and nearest reference blocks, RNN is a good type of net to predict current PU block.

5) In HEVC, there are 35 modes. It will be better if we concern each mode separately or divide these modes into several groups. But more networks will be trained and it will also require more memory and computational complexity. The ideas discussed above will provide some inspirations indicator for the future researches.

Table 2 MSE for upper left three 8x8 reconstruction blocks

| QP | Original MSE | Target MSE |
| --- | --- | --- |
| 22 | 7.73E+2 | 6.42E+2 |
| 27 | 1.61E+3 | 1.29E+3 |
| 32 | 3.23E+3 | 2.60E+3 |
| 37 | 6.48E+3 | 5.16E+3 |
| All average | 3.02E+3 | 2.42E+3 |

## 6. Conclusion

This paper proposes a CNN-based intra prediction framework. The framework is composed of training process and intra prediction process. In the training process, the CNN network learns some basic rules that the current pixels are analogous with the nearest reference pixels in one way. Besides, some latent patterns are also learned by the CNN. In the intra prediction process, the original PU block will be replaced by the target PU block achieved from the network. Experimental results show that the proposed algorithm improves the coding efficiency by 0.70% on average. This is one of the earliest attempt that applying CNN to intra prediction in HEVC. This paper can provide more inspiration for further researches.